\documentclass[sigconf]{acmart}

\usepackage{subfig}
\usepackage{caption}
\usepackage{multirow}
\usepackage{listings}
\usepackage{threeparttable}
\usepackage{arydshln}
\usepackage{xurl}

\AtBeginDocument{%
  \providecommand\BibTeX{{%
    \normalfont B\kern-0.5em{\scshape i\kern-0.25em b}\kern-0.8em\TeX}}}

\setcopyright{acmcopyright}

\copyrightyear{2022}
\acmYear{2022}
\setcopyright{rightsretained}
\acmConference[JCDL '22]{The ACM/IEEE Joint Conference on Digital Libraries in 2022}{June 20--24, 2022}{Cologne, Germany}
\acmBooktitle{The ACM/IEEE Joint Conference on Digital Libraries in 2022 (JCDL '22), June 20--24, 2022, Cologne, Germany}\acmDOI{10.1145/3529372.3530932}
\acmISBN{978-1-4503-9345-4/22/06}

\begin{document}

\title{A Domain-adaptive Pre-training Approach for Language Bias Detection in News}


\author{Jan-David Krieger}
\email{jan-david.krieger@uni-konstanz.de}
\authornote{Both authors contributed equally to this research.}
\affiliation{%
  \institution{University of Konstanz}
  \city{Konstanz}
  \country{Germany}
}

\author{Timo Spinde}
\email{timo.spinde@uni-wuppertal.de}
\authornotemark[1]
\affiliation{%
  \institution{University of Wuppertal}
  \city{Wuppertal}
  \country{Germany}
}

\author{Terry Ruas}
\email{ruas@uni-wuppertal.de}
\affiliation{%
  \institution{University of Wuppertal}
  \city{Wuppertal}
  \country{Germany}
}

\author{Juhi Kulshrestha}
\email{juhi.kulshrestha@uni-konstanz.de}
\affiliation{%
  \institution{University of Konstanz}
  \city{Konstanz}
  \country{Germany}
}

\author{Bela Gipp}
\email{gipp@cs.uni-goettingen.de}
\affiliation{%
  \institution{University Göttingen}
  \city{Göttingen}
  \country{Germany}
}

\renewcommand{\shortauthors}{Krieger, Spinde et al. }

\begin{abstract}
Media bias is a multi-faceted construct influencing individual behavior and collective
decision-making. Slanted news reporting is the result of one-sided and polarized writing which can occur in various forms. In this work, we focus on an important form of media bias, i.e. \textit{bias by word choice}. Detecting biased word choices is a challenging task due to its linguistic complexity and the lack of representative gold-standard corpora. We present DA-RoBERTa, a new state-of-the-art transformer-based model adapted to the media bias domain which identifies sentence-level bias with an F1 score of 0.814. In addition, we also train, DA-BERT and DA-BART, two more transformer models adapted to the bias domain. Our proposed domain-adapted models outperform prior bias detection approaches on the same data. 
\end{abstract}




\maketitle

\section{Introduction}\label{sec:intro}
Over the last few years, online news has increasingly replaced traditional printed news formats \cite{Dallmann2015,Kaye2016,icwsm18mediabiasmonitor, https://doi.org/10.1111/ajps.12539}. 
Online news environment provides information from diverse sources with varying perspectives, thereby, allowing people to decide which source they want to consume \cite{babaei2018purple}.
Unfortunately, the diversity of online news sources also opens the door for slanted and non-neutral news coverage \cite{10.1145/3383583.3398619}. 
Biased news coverage - referred to as \textit{media bias} in the literature \cite{Spinde2021f, spinde2021g, Spinde2021Embeddings} - occurs once subjective reporting on a specific event replaces objective coverage. 
Media bias manifests in various forms such as bias by word choice \cite{Spinde2021AutomatedIO} or bias by omission \cite{Alonso2015omission} of information. For more examples of media bias, we refer to \cite{Spinde2021AutomatedIO}.

Detecting and potentially reducing media bias in the news is societally relevant on multiple accounts. 
For policy regulators and related organizations, automated bias detection can help keep a tab on the bias in different outlets in the news ecosystem. 
For news consumers, it can help the development of tools for mitigating any adverse effect of the media bias on them. 
For journalists, automatic bias identification can improve their writing through more objective reporting \cite{Spinde2021AutomatedIO}. 
Ideally, in the future, journalistic writing tools would mitigate biases in news reporting by accurately prompting reporters when their news coverage exhibits linguistic bias.

Detecting media bias is a complex task due to its subtle nature and the lack of a clear and unique linguistic definition. 
Therefore, developing accurate quantitative detection approaches is known to be a challenging task in media bias research \cite{Spinde2021AutomatedIO,lim-etal-2020-annotating,spinde2021think,ganguly2020empirical}. 
One of the main problems in media bias research is the lack of exhaustive gold-standard bias datasets for pre-training large-scale language models (e.g., BERT \cite{devlin2019bert}). 
Prior transformer-based approaches tackle the resource limit by incorporating bias-related datasets into pre-training techniques such as \textit{Distant Supervision Learning} \cite{Spinde2021f} and \textit{Multi-task Learning} \cite{Spinde2022a} yielding performance improvements in only some experimental setups. 
Respective studies either rely on noisy and marginally bias-related training data \cite{Spinde2021f}, or do not fully exploit highly bias-related data  by incorporating only sub-samples of bias corpora into pre-training \cite{Spinde2022a}.

We propose an effective \textit{domain-adaptive pre-training} approach that relies on a highly relevant bias-related encyclopedia data set. 
Similar approaches have been shown to yield substantial performance boosts for similar tasks within the news, biomedical, and scientific domains \cite{Gururangan2020,han2019unsupervised,sun2019,lee2020biobert,beltagy2019scibert,WahleARM22}. 
To the best of our knowledge, domain-adaptive pre-training has not yet been explored in the media bias domain. 

Our primary research objective is to assess the effects of domain-adaptive pre-training on the media bias detection performance of several large-scale language models. 
Our key contribution is to leverage transformer-based models with an understanding of biased language. 
We perform an intermediate pre-training procedure with \textit{BERT} \cite{devlin2019bert}, \textit{RoBERTa} \cite{liu2019roberta}, \textit{BART} \cite{lewis2019bart}, and \textit{T5} \cite{raffel2020exploring} on the \textit{Wiki Neutrality Corpus} (WNC) \cite{pryzant2020automatically}, which contains 180k sentence pairs from \textit{Wikipedia} labeled as biased/neutral \cite{pryzant2020automatically} and fine-tune the architecture on the state-of-the art media bias data set \textit{BABE} \cite{Spinde2021f}. 
We publish our domain-adapted models, i.e. \textit{DA-RoBERTa} (DA = domain-adaptive), \textit{DA-BERT}, \textit{DA-BART}, and \textit{DA-T5}, as well as training data and all material on \url{https://github.com/Media-Bias-Group/A-Domain-adaptive-Pre-training-Approach-for-Language-BiasDetection-in-News}. DA-RoBERTa achieves a new state-of-the-art performance on BABE (F1 = 0.814), while DA-BERT, DA-BART, and DA-T5 also outperform the baselines and distantly supervised models from prior work \cite{Spinde2021f}.

\section{Related Work}\label{sec:related_work}

While media bias occurs in various forms (e.g, bias by omission, editorial bias) \cite{Spinde2021AutomatedIO}, our work focuses on bias by word choice induced by choosing different words to refer to the same concept \cite{Spinde2021AutomatedIO}. 
A detailed introduction on different media bias forms can be found in \citet{Recasens2013}. 

Several studies tackle the challenge of identifying biased language automatically. 
Early approaches used hand-crafted linguistic features to detect slanted news coverage on word- \cite{Recasens2013,Spinde2021AutomatedIO} and sentence-level \cite{Hube2018detecting} based on traditional machine learning techniques. 
Since these approaches have been shown poor performance in bias detection, we do not experiment with manually generated bias-inducing features. 
Instead, we only include feature-based results from \citet{Spinde2021f} as a baseline in our experiments. 
A detailed introduction of feature-based bias detection studies can be found in \citet{Spinde2021f} .

In the rest of the section, we first summarize drawbacks of existing media bias corpora and justify why we focus on a single state-of-the-art bias corpus for evaluative purposes. 
Next, we discuss relevant transformer-based bias detection approaches and domain-adaptive pre-training studies.

\subsection{Drawbacks of existing bias corpora}
Several approaches tackle the challenge of creating representative media bias data sets \cite{lim-etal-2020-annotating,Lim2018UnderstandingCO,baumer2015framing,Spinde2021MBIC,Spinde2021f, Spinde2021c}. 
However, most corpora exhibit substantial drawbacks such as low inter-annotator agreement \cite{lim-etal-2020-annotating,lim-etal-2020-annotating,baumer2015framing,Spinde2021MBIC,ganguly2020empirical}, low number of covered topics \cite{Lim2018UnderstandingCO}, or they focus on other concepts such as framing rather than on bias \cite{baumer2015framing}.

To the best of our knowledge, the most exhaustive media bias data set - BABE (Bias Annotations By Experts) \citet{Spinde2021f}, contains 3700 sentences covering a wide range of topics and news articles from various news outlets. 
Five media bias experts labeled sentences in terms of bias on sentence- and word-level, among others. 
The resulting inter-annotator agreement on sentence-level is 0.39 measured by Krippendorff's $\alpha$ \cite{Krippendorff2011ComputingKA}, which is much higher compared to other corpora.
Therefore, we solely rely on BABE for our experimental evaluations, since no other bias data set exhibits similar data quality and representativeness.
 We plan to conduct more experiments applying our domain-adaptive approach in future datasets, assuming they will incorporate the aspects already present in BABE.

\subsection{Transformer-based detection approaches}\label{sec:transformers}

The linguistic subtlety of slanted news coverage is known to be a great challenge for automated classification methods \cite{Spinde2021f}. 
Recent media bias studies have progressed from manually generated linguistic features \cite{10.1145/3383583.3398585, Spinde2020} to state-of-the-art NLP models yielding internal word representations by unsupervised or supervised training on massive text corpora. 
The Transformer architecture \cite{vaswani2017attention} has shown superior performance in several downstream tasks, such as, text classification \cite{OstendorffRBG20,OstendorffRSR20, OstendorffBRG22}, plagiarism detection \cite{WahleRFM22, WahleRMG21b}, word sense disambiguation \cite{WahleRMG21a} and fake news detection on the health domain \cite{WahleARM22}. 
However, the use of neural language models, such as BERT \cite{devlin2019bert} and RoBERTa \cite{liu2019roberta} in the media bias domain is still incipient \cite{Spinde2021f,Spinde2022a}. In this work, we contribute to mitigate this problem by applying the aforementioned language models via a domain-adaptive approach \cite{han2019unsupervised,sun2019,Gururangan2020}.

\citet{Spinde2021f} pre-train transformer-based models such as BERT \cite{devlin2019bert}, RoBERTa \cite{liu2019roberta}, and
DistilBERT \cite{sanh2020distilbert} using Distant Supervision Learning on news headlines from articles with different political leanings and fine-tune it on BABE \cite{Spinde2021f}. 
Their best-performing models classify biased/non-biased sentences extracted from BABE with F1 scores of 0.804 (BERT) and 0.799 (RoBERTa). 
The authors also incorporate a feature-based classifier and show that transformer models substantially outperform the feature-based approach. 
As transformer-based models have been shown to clearly outperform feature-based ones, we exclude the latter from our experiments.

\citet{Spinde2022a} train DistilBERT \cite{sanh2020distilbert} on combinations of bias-related datasets using a Multi-task Learning (MTL) \cite{WorshamK20,ChenZY21} approach. 
Their best-performing MTL model achieves 0.776 F1 score on a subset of BABE. 
However, the MTL model is outperformed by a baseline model (F1 = 0.782) trained on a subset of the datasets (WNC) used. 
\citet{Spinde2022a} suggest that improvements can be attributed to the WNC dataset being strongly bias-related, hence equipping the model with bias-specific knowledge.

While \citet{Spinde2021f} do not fully exploit bias-related data sets in their pre-training approach, \citet{Spinde2022a} implement a complex MTL architecture reducing the WNC's pre-training effect on the bias classification task.
In our work, we use a similar learning task as \citet{Spinde2021f} and exploit the WNC's bias-relatedness by extending the pre-training of several transformer models on the whole WNC instead of its subset. 

\subsection{Domain-adaptive pre-training approaches}

Our training setup can be considered a form of domain-adaptive pre-training \cite{han2019unsupervised,lee2020biobert,beltagy2019scibert} in which a language model is equipped with domain-specific knowledge. 
Several studies experiment with domain-adaptive learning approaches in different domains (e.g., BioBERT \cite{lee2020biobert}, SciBERT \cite{beltagy2019scibert}), but none of them deals with media bias detection \cite{han2019unsupervised,lee2020biobert,beltagy2019scibert,sun2019,Gururangan2020}. 

\citet{sun2019} explore different techniques for domain-adaptive pre-training of BERT for text classification tasks such as sentiment classification, question classification, and topic classification.
BERT is additionally pre-trained on data from various domains leading to performance boosts on many tasks if the training data are related to the target task’s domain. 
After training BERT on several sentiment classification datasets,  \citet{sun2019} reduced the error rate on the Yelp sentiment data set to 1.87\% (compared to 2.28\% from BERT baseline initialized with \textit{bert-base-uncased} weights).
The results from \citet{sun2019} are supported by \citet{Gururangan2020} investigating domain-adaptive pre-training of RoBERTa in four different target domains (i.e., biomedical, computer science publications, news, and reviews) and eight subsequent classification tasks. 
When pre-training RoBERTa on large amounts of news text, the model's F1-score on a hyperpartisan classification dataset \cite{kiesel-etal-2019-semeval} improves from F1 = 0.886 (\textit{roberta-base} weights) to F1 = 0.882. Training the model on a domain outside the domain of interest (\textit{irrelevant domain-adaptive pre-training}) drastically decreases performance to F1 = 0.764.

Our domain-adaptive pre-training approach is performed on the WNC corpus and based on implementations from \citet{sun2019} and \citet{Gururangan2020}. Due to drastic performance increases through irrelevant domain-adaptive pre-training in previous research \cite{Gururangan2020}, we do not implement respective experiments.
We detail our proposed training process and experiments in Sections \ref{sec:methods} and \ref{sec:exp}. 
Since most existing approaches focus on sentence-level bias detection, we follow the standard practice and develop a sentence-level classification model. 
Compared to cutting-edge but convoluted studies in media bias detection \cite{Spinde2021d,Spinde2021f}, we perform a more focused and direct training setup on a large amount of highly bias-related data and expect substantial performance improvements.

\section{Methodology}\label{sec:methods}
\begin{figure}
    \centering
    \includegraphics[width = 0.4\textwidth]{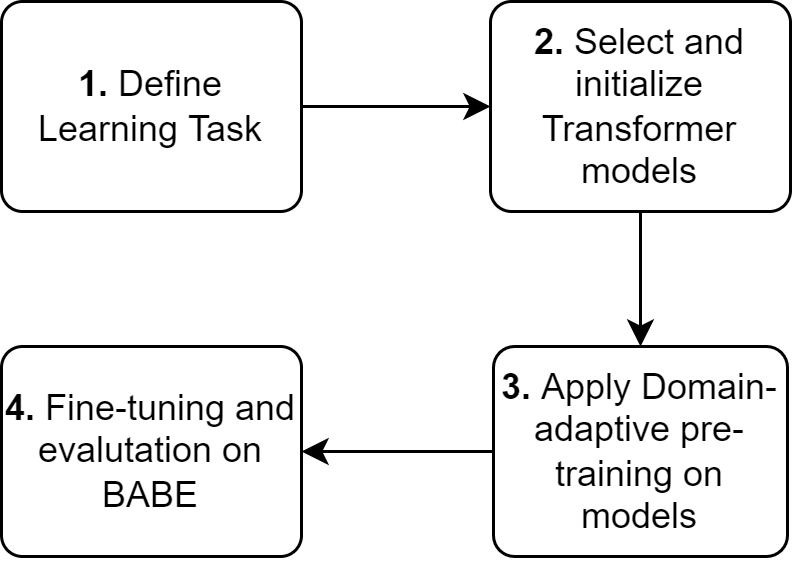}
    \caption{Pre-training and fine-tuning workflow.}
    \label{fig:pipeline}
\end{figure}

We use neural-based language models, pre-train them on the bias domain (WNC), and perform evaluations on the media bias classification task using BABE as Figure \ref{fig:pipeline} shows.
We expect that domain-adaptive pre-training improves word representations by adapting them to the data distributions of biased and non-biased news content. 
Based on BABE, we define a learning task that is later optimized (Section \ref{sec:learning_task}). 
Then, we select suitable transformer models and initialize them with pre-trained weights (section \ref{sec:transformers-method}). 
We adapt the models to the media bias domain by training them on the WNC  (section \ref{sec:domain-adaption}).
Finally, all models are fine-tuned and evaluated on BABE.

\subsection{Learning task}\label{sec:learning_task}

The language models are optimized via intermediate training. We have a corpus $X$ containing sentences $x_i \in X$ with $i = {1,...,N}$ and binary bias labels (\textit{Biased} vs. \textit{Non-biased}) encoded as $1$ and $0$, respectively.  The task is to assign the correct label $y_i \in \{0,1\}$ to $x_i$. The training objective is to minimize a binary cross-entropy loss 

\begin{equation}
\label{eq:loss}
\mathcal{L} := - \frac{1}{N} \sum_{i=1}^{N} \sum_{k=\{0,1\}} f_k(x_i) \cdot log(\hat{f_k}(x_i)).
\end{equation}

where $f_k(x_i)$ refers to the true binary label and $\hat{f_k}(x_i)$ indicates the model's predicted score for a sentence.

\subsection{Transformer-based models} \label{sec:transformers-method} 
We choose BERT and RoBERTa for our domain-adaptive pre-training as they represent the best-performing models in \citet{Spinde2021f}. 
Doing so, we also achieve maximum comparability to previous state-of-the-art bias classifiers. 
Additionally, we incorporate BART and T5, since encoder-decoder architectures have demonstrated a clear improvement in comparison to BERT in several NLP tasks (e.g., GLUE \cite{wang-etal-2018-glue}).
We choose the corresponding models to investigate how the combination of autoencoder and autoregressive components (BART), and advanced MTL architectures (T5) perform on our media bias detection task.

BERT learns bidirectional word representations on unlabeled text optimizing an unsupervised learning task based on \textit{Masked Language Modeling} and \textit{Next Sentence Prediction}. 
In contrast to BERT, RoBERTa drops the Next Sentence Prediction task and differs slightly in terms of pre-training data.
BART uses text manipulations by noising and learns representations by reconstructing the original text sequence. T5 uses an MTL architecture pre-trained on various supervised and unsupervised tasks by converting all training objectives into text-to-text tasks.
All models are adapted to the media bias domain (Section \ref{sec:domain-adaption}) and evaluated on the sentence-level media bias classification task (Section \ref{sec:exp}).

\subsection{Domain-adaptive pre-training}\label{sec:domain-adaption}
Adapting a pre-trained language model to a specific domain becomes essential when the target domain differs strongly from the pre-training ground truth \cite{han2019unsupervised,lee2020biobert,beltagy2019scibert}. 
Due to tendentious and dubious vocabulary in slanted news, media bias is different from most of the domains BERT-like models are pre-trained on. 
For example, BERT is trained on English Wikipedia and the BooksCorpus \cite{zhu2015aligning} while RoBERTa additionally incorporates commonsense reasoning data, news data, and web text data. 
To the best of our knowledge, a specific BERT-like model trained on biased language in news does not exist to date. 
BERT models pre-trained on fake news \cite{blackledge2021transforming} and political orientation classification \cite{gupta2020polibert} do exist. 
However, the concepts of fake news and political orientation differ substantially from the media bias domain. 

Our domain-adaptive pre-training uses the WNC to optimize our learning task defined in Section \ref{sec:learning_task}. 
The 180k sentence pairs contained in the corpus are manually selected from Wikipedia articles as going against the platform’s \textit{Neutral Point of View} (NPOV) standard\footnote{\url{https://en.wikipedia.org/wiki/Wikipedia:Neutral_point_of_view}}. 
The pairs contain an original biased sentence and its manually derived neutral counterpart. 
Bias forms included in the corpus refer to \textit{epistemological bias}, \textit{framing bias}, and \textit{demographic bias}. 
\citet{Recasens2013} define framing bias as choosing subjective words to embed a particular point of view in the text whereas epistemological bias is described as a modification of a statement's plausibility.
\citet{pryzant2020automatically} introduce demographic bias as text containing predispositions towards a certain gender, race, or other demographic category. 
For a detailed description on sentence selection criteria and the revision process, see \citet{pryzant2020automatically}. 

Our approach is inspired by \citet{sun2019} and \citet{Gururangan2020}, which conclude domain-adaptive pre-training is most efficient once pre-training data for the domain adaption is related to the target domain and task. 
Since WNC (pre-training) and BABE (fine-tuning) have similar bias forms, and are both composed of manually labeled sentences (biased and non-biased), we expect the proposed pre-training task to improve our fine-tuning results.

\subsection{Experiments}\label{sec:exp}

\subsubsection{Pre-training}
We initialize RoBERTa, BERT, BART, and T5 with pre-trained weights provided by the HuggingFace API\footnote{\url{https://huggingface.co/}}, and stack a dropout layer (Dropout = 0.2) and randomly initialized linear transformation layer (768,2) on top of the model. 
All models are used in their base form.

For the domain-adaptive pre-training, sentences are batched together with 32 sentences per batch. 
For model optimization, we use the AdamW optimizer\footnote{\url{https://huggingface.co/docs/transformers/main_classes/optimizer_schedules}} with a learning rate of $1e^{-5}$, and model performance is evaluated on binary cross-entropy loss. Model convergence can be observed after one epoch and a runtime of $\approx5$ hours on a Tesla P100-PCIE GPU with 16GB RAM.

\subsubsection{Fine-tuning}
We fine-tune and evaluate the model on BABE \citet{Spinde2021f} with a
batch size = 32. 
We again use the AdamW optimizer (learning rate = $1e^{-5}$), and model convergence based on cross-entropy loss can be observed after 3-4 epochs. 
Due to the small data size of 3700 sentences, we report the model’s F1 score in the binary bias labeling task averaged by 5-fold cross-validation. 
Fine-tuning is performed on a Tesla K80 GPU (12GB RAM) in  $\approx15$ minutes.

\subsubsection{Baseline}
For every domain-adaptive language model, we compare its sentence classification performance on BABE to the same architecture merely fine-tuned on BABE (without domain-adaptive pre-training as an intermediate training step).
Thereby, we can assess the effect of our training approach.
Since \citet{Spinde2021f} achieve state-of-the-art results on BABE with Distant Supervision Learning \cite{Spinde2021f}, we additionally compare our F1 scores to their scores achieved by training BERT and RoBERTa on news headlines distantly labeled as biased and non-biased. 
We provide statistical significance tests for our domain-adapted models vs. fine-tuned-only models.

\subsubsection{Test for Statistical Significance}
In their review on existing NLP studies, \citet{dror-etal-2018-hitchhikers} report that most approaches lack statistical tests inspecting the significance of experimental results. 
The authors recommend various parametric and non-parametric test to compare performances of Machine Learning models. 

For our approach, we select the \textit{McNemar\'s test} which is a non-parametric test to compare the performance of two algorithms on a target task. 
Since we do not have information on the distribution of our target metric (F1 score), a non-parametric approach is a suitable option to test for significance. 
The test is based  on a $2x2$ contingency table showing the models' predictions on $n$ instances of a target task\'s test set. 
Under the null hypothesis $H_0$, the test assumes that both algorithms output the correct/incorrect label for the same proportion of instances from the test set. 
Accordingly, the alternative hypothesis $H_1$ states that both algorithms differ significantly in terms of their agreement on items from the test set. 
The test statistic follows a $\chi^2$ distribution and is suitable for NLP tasks such as binary text classification \cite{dror-etal-2018-hitchhikers,dietterich1998approximate}. 
For a more detailed introduction on statistical significance tests for NLP use cases, see \citet{dror-etal-2018-hitchhikers}.

\section{Results}
Table \ref{Tab:class_res} shows the F1 scores (averaged over 5-fold CV split) of our transformer-based experiments on the binary sentence classification task. 
All domain-adapted models (third block) outperform the baselines models (first block) and the distantly supervised models (middle block) trained by \citet{Spinde2021f}. 

The best-performing model that achieves a new state-of-the-art on BABE is DA-RoBERTa (F1 = 0.814), surpasses the baselines and its Distant Supervision variant by 1.5 \%. 
DA-BERT, DA-BART, and DA-T5  achieve a lower F1-score of 0.809, 0.809, and 0.798, yet outperform BERT, BART, and T5 by 2\%, 0.8\%, and 1.2\%, respectively. 
However, DA-BERT increases sentence classification performance by only 0.5\% compared to BERT trained via Distant Supervision \cite{Spinde2021f}. 
To the best of our knowledge, a distantly supervised variant for BART and T5 is not available.

\begin{table}[ht!]
\caption{Stratified 5 fold cross-validation results.}
\centering
\resizebox{0.45\textwidth}{!}{
\begin{threeparttable}
\begin{tabular}{p{4cm}c}
\hline
\textbf{Model} & \textbf{Macro F1 (error)} \\
\hline
BERT & 0.789 (0.011)  \\
RoBERTa & 0.799 (0.011)   \\
BART & 0.801 (0.009)  \\
T5 & 0.786 (0.008)  \\
\cdashline{1-2}
BERT-distant \cite{Spinde2021f} & 0.804 (0.014) \\
RoBERTa-distant \cite{Spinde2021f}  & 0.799 (0.017) \\
\cdashline{1-2}
DA-BERT & 0.809 (0.010) \\
DA-RoBERTa & \textbf{0.814} (0.004) \\
DA-BART & 0.809 (0.009)  \\
DA-T5 & 0.798 (0.009)  \\
\hline
\end{tabular}
\begin{tablenotes}
\small
\item \textit{Note:} Standard errors across folds in parentheses. 
\item The first block shows results of baseline approaches without intermediate pre-training. The second block shows results from \cite{Spinde2021f} based on Distant Supervision Learning (BART and T5 are not incorporated in their study). Results from our domain-adaptive approach are shown in the third block.
\item The best result is printed in \textbf{bold}.
\end{tablenotes}
\end{threeparttable}
}

\label{Tab:class_res}
\end{table}

Table \ref{Tab:mcnemar}  shows results of the McNemar statistical significance tests comparing our domain-adapted models with respective baselines. 
On a significance level of $\alpha = 0.05$, we can observe significant F1-score improvements for BERT vs. its domain-adapted variant ($\chi^2 = 5.65, p = 0.031$) as well as for RoBERTa vs. DA-RoBERTa ($\chi^2 = 3.844, p = 0.049$) and T5 vs. DA-T5 ($\chi^2 = 4.86, p = 0.027$). 
Adapting BART to the bias domain seems not to significantly improve the sentence classification performance ($\chi^2 = 3.629, p = 0.057$).

\begin{table}[ht]
\caption{Results of the McNemar test for statistical significance between baseline (without domain-adaptive pretraining) and domain-adapted models.}

\centering
\resizebox{0.45\textwidth}{!}{
\begin{threeparttable}
\begin{tabular}{p{4cm}cc}
\hline
\multirow{2}{*}{\textbf{Models}}  & \multicolumn{2}{c}{\textbf{McNemar test statistic}}  \\
 \cline{2-3} 
 & $\chi^2$ & $p$ \\
\hline
BERT vs. DA-BERT & 5.65 & 0.031*    \\
RoBERTa vs. DA-RoBERTa & 3.84 & 0.049*    \\
BART vs. DA-BART & 3.63 & 0.057  \\
T5 vs. DA-T5 & 4.86 & 0.027*  \\
\hline
\end{tabular}
\begin{tablenotes}
\small
\item \textit{Note:} $*p<.05$
\end{tablenotes}
\end{threeparttable}
}
\label{Tab:mcnemar}
\end{table}

\section{Discussion}

With DA-RoBERTa, we provide a new state-of-the-art classifier for the detection of biased language in the news articles on sentence-level. 
Furthermore, we show that all domain-adapted models outperform their baselines and distantly supervised models published by \citet{Spinde2021f}. 
Our results can be considered a contribution towards a sufficiently accurate bias detection tool. 
However, some significance tests comparing the performance of domain-adapted models vs. distantly supervised models are missing due to limited resources.

As indicated in Section \ref{sec:transformers}, \citet{Spinde2022a} pre-train DistilBERT on a subset of the WNC and observe performance boosts of 3.6\% on bias sentence classification with a subset of BABE compared to their baseline without intermediate pre-training. 
Although our domain-adapted models incorporate the complete WNC into pre-training, we observe minor performance increases when compared to those obtained by DistilBERT. 
We believe that smaller-scaled and distilled models such as DistilBERT benefit more from additional pre-training than larger models relying on a different training objective such as BERT, RoBERTa, BART, and T5. 

In the future, it will be interesting to verify how even more robust and general NLP models benefit from intermediate pre-training. 
Possibly, state-of-the art NLP models such as the recently published \textit{ExT5} \cite{aribandi2021ext5}, incorporating extensive Multi-task Learning on 107 tasks from different domains, further decreases domain-adaptive learning effects. 
Furthermore, we expect that bias corpora such as BABE will continue to be proposed. 
From a resource consumption perspective, fine-tuning robust language models such as ExT5 on more representative bias corpora might be sufficient to achieve state-of-the-art performances in bias detection.

Considering our bias detection task, we want to point out that our models are merely trained to identify slanted news coverage on sentence-level. 
Since media bias is a linguistically complex construct \cite{Spinde2021AutomatedIO}, we need robust and more general classifiers for different linguistic bias perspectives such as word-, paragraph-, and article-level. 
As \citet{Recasens2013} show, word-level detection of slanted news coverage is challenging for both humans and machines.
Computer Science approaches dealing with bias on word-level might depend on collaborations with researchers from the Social Sciences to develop a large number of linguistically fine-grained gold-standard data for efficient model training. 
Furthermore, we need systems detecting various sub-forms of bias such as framing bias and epistemological bias accurately. 
MTL approaches trained on different bias categories might be a a promising direction for future models.

Future research should incorporate a broader range of evaluation tasks to assess model performance. 
\citet{Spinde2021f} argue that standard metrics such as F1 score are not sufficient to evaluate language models on the complex bias detection task. 
The authors suggest developing more advanced evaluation metrics such as decomposing the bias detection task into several subtasks to assess a model's detection power properly. 
\citet{ribeiro-etal-2020-beyond} introduce CheckList, a tool structuring the target task into several sub-tasks. Respective evaluation approaches could help assess a model's bias identification performance on different forms of bias.


\section{Conclusion}
This work proposes DA-RoBERTa, a new state-of-the-art language model for sentence-level detection of biased language in the news. 
We equip several transformer architectures (i.e., BERT, RoBERTa, BART, and T5) with an understanding of biased language, showing that domain-adaptive pre-training significantly improves the classifier's bias detection performance compared to baseline models without intermediate pre-training. 
Limitations of our approach are the exclusively pre-training focus on sentence-level classification and the restricted evaluation incorporating a single data set/task due to the lack of existing representative bias corpora. We hope that further high-quality bias corpora are published in the future to improve the generalizability of results and enable a more fine-grained and large-scale evaluation of models in the domain. 
Considering continuous developments in the NLP field, future studies should also address whether upcoming more robust language models still require intermediate pre-training on the media bias domain.

\section*{Acknowledgments}

The Hanns-Seidel-Foundation, Germany, supported this work, as did the DAAD (German Academic Exchange Service).

\bibliographystyle{ACM-Reference-Format}
\bibliography{references}


\end{document}